\documentclass[11pt]{article}

\usepackage{acl}

\usepackage{times}
\usepackage{latexsym}
\usepackage[T1]{fontenc}
\usepackage[utf8]{inputenc}
\usepackage{microtype}
\usepackage{inconsolata}
\usepackage{graphicx}
\usepackage{booktabs}
\usepackage{enumitem}
\usepackage{amsmath}
\usepackage{multirow}
\usepackage{subcaption}
\usepackage{pifont}
\newcommand{\cmark}{\ding{51}}
\newcommand{\xmark}{\ding{55}}

\newcommand{\eqcontribnote}[1]{%
  \begingroup
    \renewcommand\thefootnote{}\footnote{#1}%
    \addtocounter{footnote}{-1}%
  \endgroup
}

\title{Faithfulness Is Not Free: Auditing Offline {KV}-Cache Quantization\\
       in Retrieval-Augmented Generation}

\author{
  Atta Ul Asad\textsuperscript{1*} \quad
  Ahsan Bilal\textsuperscript{2*} \quad
  Muhammad Ali\textsuperscript{3} \quad
  Muhammad Haseeb\textsuperscript{4} \quad
  Dean F. Hougen\textsuperscript{2} \\[4pt]
  \textsuperscript{1}Lahore University of Management Sciences (LUMS) \quad
  \textsuperscript{2}University of Oklahoma \\
  \textsuperscript{3}Air University \quad
  \textsuperscript{4}National University of Sciences and Technology (NUST) \\[3pt]
  \texttt{26280099@lums.edu.pk} \quad
  \texttt{\{ahsan.bilal-1, hougen\}@ou.edu}
}

\begin{document}
\maketitle
\eqcontribnote{$^{*}$\,Equal contribution (core contributors).}

\begin{abstract}
Retrieval-augmented generation systems can precompute and store key-value caches of retrieved documents to avoid re-encoding context at every query. Quantizing these caches further reduces storage, but no prior work asks whether compression damages \emph{faithfulness}, whether responses remain grounded in the retrieved evidence. Faithfulness and accuracy are not equivalent: a model can produce a correct answer that is no longer supported by the context it was given. We evaluate Qwen2.5-7B-Instruct under INT8 and INT4 quantization on RGB and HotpotQA, measuring both accuracy and faithfulness with a hallucination detector, NLI entailment, and an LLM judge. INT8 is near-lossless across both metrics. INT4 reduces accuracy and, more critically, even among answers that remain factually correct, over 90\% of faithfulness changes are negative, i.e., accuracy metrics are blind to this regression. The harm grows under noisy retrieval and with more retrieved chunks. Faithfulness must be audited before compressed caches are deployed.
\end{abstract}

\section{Introduction}
Retrieval-augmented generation (RAG) grounds model responses in external documents, but re-processing the retrieved context for every query is expensive. Recent systems eliminate this cost by precomputing key-value (KV) caches for documents once and reusing them at inference time \citep{turborag_2024,cacheblend_2025}. As these offline caches scale with corpus size and retrieval depth, their storage footprint becomes a bottleneck, making low-bit quantization an attractive solution.

Prior work shows KV caches can be quantized to INT8 and INT4 with little \emph{accuracy}
degradation \citep{kivi_2024,kvquant_2024}, though aggressive compression can harm
reasoning on complex tasks \citep{quant_reasoning_2025}.
However, all existing evaluations measure only task accuracy.
They leave open a more subtle question: do quantized caches preserve
\textbf{faithfulness}, whether generated responses remain \emph{supported by the
retrieved evidence}?
This distinction matters.
Exact Match and F1 measure agreement with a reference answer; they do not verify
whether the answer is grounded in the provided context.
A model can preserve the correct answer string while becoming less dependent on
retrieved evidence, creating a hidden failure mode in compressed RAG
systems \citep{helmet_2025,tamber_faithrag_2025}.

\begin{figure*}[t]
  \centering
  \includegraphics[width=\textwidth]{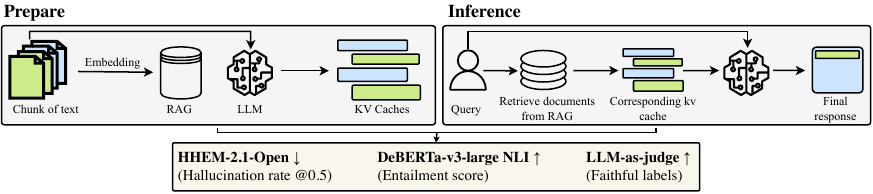}
  \caption{Audit pipeline. \textbf{Prepare}: retrieved documents are prefilled in one
  causal pass, and the resulting KV cache is quantized and stored.
  \textbf{Inference}: the stored cache is dequantized and reused for each query.
  \textbf{Evaluate}: three complementary faithfulness signals (HHEM hallucination
  rate $\downarrow$, DeBERTa-v3 NLI entailment $\uparrow$, LLM-as-judge $\uparrow$)
  are measured alongside standard accuracy metrics.}
  \label{fig:pipeline}
\end{figure*}

To our knowledge, this is the first audit of offline KV-cache quantization in RAG that evaluates faithfulness, showing that cache precision can affect grounding even when answer accuracy is preserved. Figure~\ref{fig:pipeline} shows our controlled protocol, which isolates cache precision as the sole variable and evaluates faithfulness with three independent signals. Our contributions are threefold. First, we introduce a faithfulness-centered audit protocol that avoids positional-stitching confounds by caching the full retrieved context in a single causal pass. Second, we show that INT8 is near-lossless, while INT4 silently degrades grounding even on accuracy-preserved examples ($>$90\% of faithfulness flips are negative; McNemar $p<10^{-20}$). Third, we show that this degradation is amplified by retrieval noise and larger retrieval depth, establishing faithfulness (not accuracy alone) as the necessary audit signal for compressed RAG.

\section{Background and Related Work}

\subsection{KV-cache compression.}
KIVI \citep{kivi_2024} achieves near-lossless 2-bit asymmetric quantization;
KVQuant \citep{kvquant_2024} reaches sub-4-bit via per-channel key quantization;
ZipCache \citep{zipcache_2024} identifies token-level precision sensitivity;
KVTuner \citep{kvtuner_2025} sets 4-bit as the safe floor for Qwen2.5-7B on
reasoning tasks.
Sub-4-bit can cause up to 59\% accuracy loss on long-context
tasks \citep{mekala_quant_longcontext_2025}, prompting variance-normalized
quantization \citep{kvarn_2026} and query-aware mixed-precision
allocation \citep{mixkvq_2026}.
\textbf{Gap}: all of these evaluate \emph{online} cache compression via accuracy or
perplexity; none audits faithfulness in retrieval-augmented settings.

\subsection{RAG faithfulness.}
RAGChecker \citep{ragchecker_ru_2024} and RAGTruth \citep{ragtruth_wu_2024} show that EM and F1 miss systematic faithfulness failures. HELMET \citep{helmet_2025} finds standard accuracy proxies correlate poorly with downstream RAG quality. LLM-as-judge outperforms detector-only methods for faithfulness evaluation \citep{tamber_faithrag_2025,song_trustscore_2025}. Distractor passages cause systematic grounding failures even without compression \citep{wu_ragnoise_2025,amiraz_distracting_2025}. \textbf{Gap}: none of these studies examines whether \emph{cache compression} introduces faithfulness failures beyond those from retrieval noise alone.

\subsection{Offline KV-cache RAG.}
TurboRAG \citep{turborag_2024} stores chunk-level caches offline to eliminate repeated prefill; CacheBlend \citep{cacheblend_2025} addresses per-chunk positional mismatch via selective recomputation, reducing TTFT by 2.2--3.3$\times$. \textbf{Gap}: neither examines whether compressing stored caches harms faithfulness.

\section{Method} \label{sec:method}
Our goal is to isolate the effect of KV-cache quantization in RAG. We keep the retrieved evidence, prompt, and decoding fixed, and vary only the stored cache precision. The method builds one position-consistent cache for the full retrieved context, quantizes it before storage, dequantizes it before decoding, and compares BF16, INT8, and INT4 under identical inference conditions.

\textbf{Building a unified cache.}
For each query, we concatenate the system prompt with the top-$K$ retrieved
chunks and run one causal prefill pass. The resulting KV cache is a single
unified cache for the full retrieved context, and this is the cache we compress.
We avoid caching chunks independently and stitching them later, because
independent chunk caches are created at local token positions but consumed at
global positions during generation. This positional mismatch can degrade output
quality by itself. Using a single prefill pass removes this confound and ensures
that cache bit-width is the only variable changed across conditions.

\textbf{Quantizing the cache.}
After prefill, we quantize the KV cache before storage and dequantize it back to
bfloat16 before decoding. This matches an offline RAG deployment in which
document caches are written to disk once and reused at inference time. INT8 uses
per-token asymmetric quantization \citep{kivi_2024}. INT4 uses group-wise
quantization with groups of 64 channels, where each group stores its own scale
and zero-point. This reduces the risk that large-magnitude channels dominate the
quantization range and collapse smaller values under INT4's 16 levels
\citep{kivi_2024,kvquant_2024}. The on-disk footprint of one KV cache is
\begin{equation}
M(b,g,o) =
2LHd_hN\cdot\tfrac{b}{8}
+
\tfrac{2LHd_hN}{g}\cdot\tfrac{o}{8},
\label{eq:kvsize}
\end{equation}
where $L$, $H$, $d_h$, and $N$ denote the number of layers, KV heads, head
dimension, and cached tokens. The quantization bit-width is $b$, the group size
is $g$, and $o$ is the scale/zero-point overhead per group. BF16
($b{=}16$) has no overhead term. INT8 uses $b{=}8$ and $g{=}d_h$, while INT4 uses
$b{=}4$ and $g{=}64$. Because INT8 and INT4 store metadata for each group, their
realized compression ratios are lower than the nominal $2{\times}$ and
$4{\times}$ limits.

\textbf{Precision comparisons.} 
Using the same retrieved chunks and greedy decoding, we compare four cache settings. \textbf{C0 (Oracle)} runs standard full-context generation without a cache round trip. \textbf{C1 (BF16)} stores and reloads the uncompressed cache, serving as the cache baseline. \textbf{C2 (INT8)} and \textbf{C3 (INT4)} store quantized caches and dequantize them before decoding. Thus, C1--C2 isolates the effect of INT8 compression, while C1--C3 isolates the effect of aggressive INT4 compression.

\section{Experimental Setup}\label{sec:setup}

We evaluate Qwen2.5-7B-Instruct \citep{qwen25_2024}, a bfloat16-native 7B
instruction-tuned decoder with a 32K-token context window.
Its native bfloat16 precision avoids the float16 dynamic-range overflow that silently
corrupts cached KV values on this architecture.

We use two complementary benchmarks.
\textbf{RGB} \citep{rgb_chen2024} (300 examples) provides its own positive and
distractor passages, keeping retrieval recall high and the noise structure controlled.
\textbf{HotpotQA} \citep{hotpotqa_yang2018} (300 examples, distractor split)
requires combining evidence across at least two supporting passages mixed with
distractors, testing multi-hop reasoning.
Each example is evaluated at $K\in\{1,3,5\}$ retrieved chunks,
giving $300\times3\times4=3{,}600$ generations per dataset.
Retrieval uses a FAISS index \citep{faiss_2021} over bge-small-en-v1.5 \citep{bge_small_2023}
embeddings; we retrieve the top-10 chunks and slice to the first $K$.

We report both answer accuracy and evidence faithfulness. Accuracy is measured by containment exact match (EM), which marks a prediction correct if any normalized gold alias appears in the generated answer, and by token-level F1 as a secondary overlap metric. Faithfulness is measured with three independent signals: HHEM-2.1-Open \citep{hhem_vectara2024}, reported as hallucination rate at threshold 0.5 ($\downarrow$); DeBERTa-v3-large NLI entailment \citep{deberta_nli} ($\uparrow$); and an LLM judge using Claude Haiku~4.5 ($\uparrow$). Since HHEM and NLI agree only moderately on RGB (Pearson $r{=}0.44$), we treat them as complementary diagnostics. The LLM judge achieves 92\% agreement on 50 hand-checked labels and serves as the primary faithfulness signal. We test three hypotheses using C1 BF16 as the uncompressed reference and C3 INT4 as the aggressive compression condition. For \textbf{H1}, we test whether faithfulness degrades even when accuracy is preserved. We restrict to examples whose containment-EM outcome is unchanged between C1 and C3, then apply a paired McNemar test to faithfulness label flips. For \textbf{H2}, we test whether harm increases with retrieval depth by fitting a linear slope over $K\in\{1,3,5\}$. Because only three retrieval depths are available, this test is underpowered ($p{=}0.12$) and is reported as directional. For \textbf{H3}, we test whether retrieval noise amplifies INT4 degradation by regressing the per-example INT4 faithfulness gap on distractor fraction within each $K$. H3 is evaluated only on RGB, where per-example noise ratios are available.

\begin{figure*}[t]
  \centering
  \includegraphics[width=0.77\textwidth]{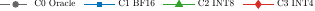}\\[3pt]

  \begin{subfigure}[t]{0.245\textwidth}\centering
    \includegraphics[width=\linewidth]{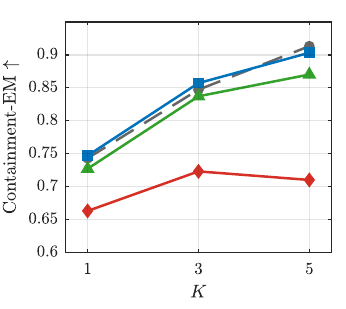}
    \caption{RGB EM accuracy}\label{fig:rgb-em}
  \end{subfigure}\hfill
  \begin{subfigure}[t]{0.245\textwidth}\centering
    \includegraphics[width=\linewidth]{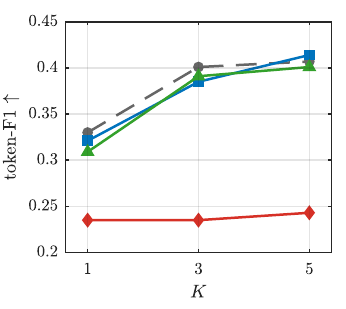}
    \caption{RGB F1}\label{fig:rgb-f1}
  \end{subfigure}\hfill
  \begin{subfigure}[t]{0.245\textwidth}\centering
    \includegraphics[width=\linewidth]{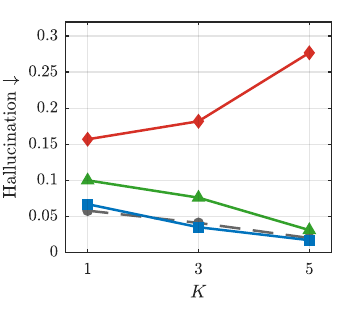}
    \caption{RGB hallucination}\label{fig:rgb-hall}
  \end{subfigure}\hfill
  \begin{subfigure}[t]{0.245\textwidth}\centering
    \includegraphics[width=\linewidth]{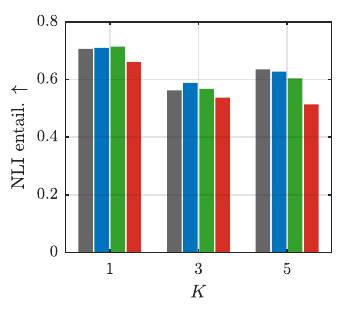}
    \caption{RGB entailment}\label{fig:rgb-entail}
  \end{subfigure}

  \begin{subfigure}[t]{0.245\textwidth}\centering
    \includegraphics[width=\linewidth]{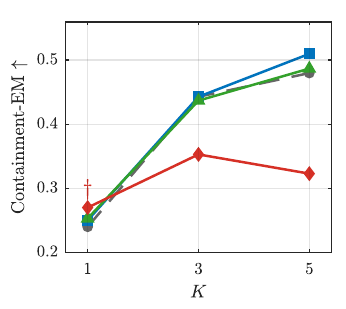}
    \caption{HotpotQA EM accuracy}\label{fig:hp-em}
  \end{subfigure}\hfill
  \begin{subfigure}[t]{0.245\textwidth}\centering
    \includegraphics[width=\linewidth]{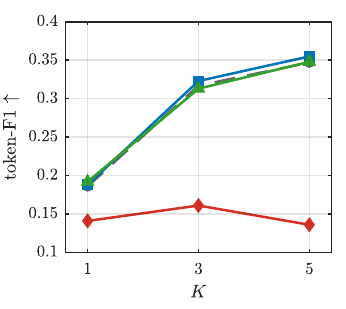}
    \caption{HotpotQA F1}\label{fig:hp-f1}
  \end{subfigure}\hfill
  \begin{subfigure}[t]{0.245\textwidth}\centering
    \includegraphics[width=\linewidth]{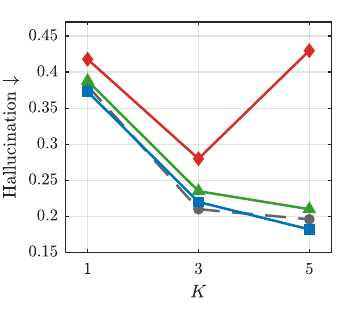}
    \caption{HotpotQA hallucination}\label{fig:hp-hall}
  \end{subfigure}\hfill
  \begin{subfigure}[t]{0.245\textwidth}\centering
    \includegraphics[width=\linewidth]{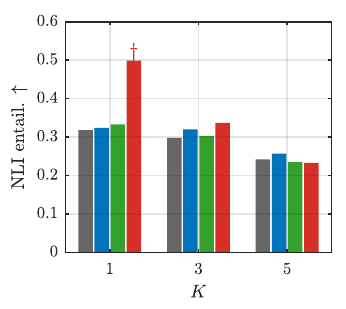}
    \caption{HotpotQA entailment}\label{fig:hp-entail}
  \end{subfigure}

  \begin{subfigure}[t]{0.245\textwidth}\centering
    \includegraphics[width=\linewidth]{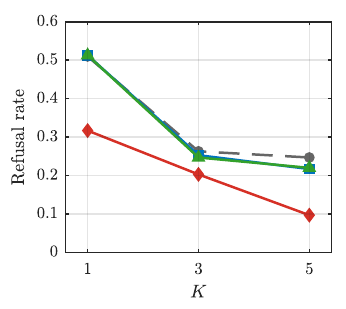}
    \caption{HotpotQA refusal}\label{fig:hp-refusal}
  \end{subfigure}\hfill
  \begin{subfigure}[t]{0.245\textwidth}\centering
    \includegraphics[width=\linewidth]{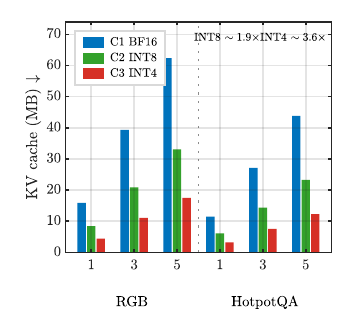}
    \caption{KV cache size}\label{fig:kv}
  \end{subfigure}\hfill
  \begin{subfigure}[t]{0.245\textwidth}\centering
    \includegraphics[width=\linewidth]{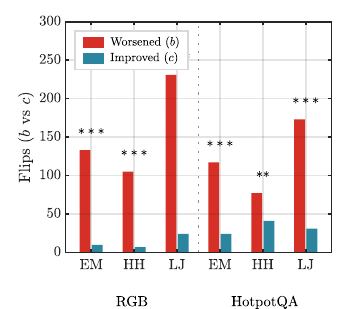}
    \caption{Hidden faithfulness harm (H1)}\label{fig:mcnemar}
  \end{subfigure}\hfill
  \begin{subfigure}[t]{0.245\textwidth}\centering
    \includegraphics[width=\linewidth]{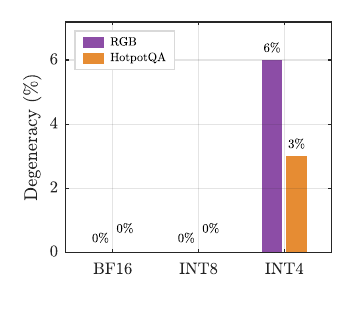}
    \caption{Degeneracy ($K{=}5$)}\label{fig:degen}
  \end{subfigure}
\caption{Benchmark results across RGB and HotpotQA under four precision conditions: C0--C3. $\dagger$ marks the HotpotQA $K{=}1$ refusal artifact: INT4 under-refuses, which
\emph{inflates} surface EM and NLI entailment---not a genuine gain.
In~(\subref{fig:kv}), tick labels are \emph{dataset,K} (R${=}$RGB, H${=}$HotpotQA);
the inset shows on-disk shrinkage vs.\ C1 BF16: $\sim$1.9$\times$ (INT8) and
$\sim$3.6$\times$ (INT4) at $K{=}5$, below the nominal 4$\times$ because INT4 stores
per-block scale/zero-point pairs.
In~(\subref{fig:mcnemar}), columns group by dataset; within each, EM${=}$containment
exact-match, HH${=}$HHEM, LJ${=}$LLM-judge; bars are McNemar flip counts
$b$ (worsened) vs.\ $c$ (improved).
Significance: ${\ast\ast\ast}\,p{<}10^{-3}$, ${\ast\ast}\,p{<}10^{-2}$,
${\ast}\,p{<}0.05$.}
  \label{fig:all-results}
\end{figure*}
\section{Results} \label{sec:results}
Figure~\ref{fig:all-results} reports the complete results across RGB and HotpotQA, and Table~\ref{tab:summary} summarizes the main effects. 

\textbf{Accuracy and hidden faithfulness harm.} INT8 remains close to the BF16 cache baseline across accuracy and faithfulness metrics, whereas INT4 consistently degrades performance. INT4 reduces containment-EM (as shown in Figs.~\ref{fig:rgb-em},~\ref{fig:hp-em}) and token-F1 (as shown in Figs.~\ref{fig:rgb-f1},~\ref{fig:hp-f1}); correct$\to$wrong flips greatly outnumber wrong$\to$correct flips on both RGB ($133$ vs.\ $10$) and HotpotQA ($117$ vs.\ $24$; Fig.~\ref{fig:mcnemar}). More importantly, INT4 also causes faithfulness loss that accuracy does not expose. On the accuracy-preserved subset, where containment-EM is unchanged between C1 and C3, faithfulness flips remain overwhelmingly negative: the LLM judge records $231$ worsenings vs.\ $24$ improvements on RGB ($p{<}10^{-38}$) and $173$ vs.\ $31$ on HotpotQA ($p{<}10^{-30}$; Fig.~\ref{fig:mcnemar}). HHEM corroborates this trend on RGB for all $K$ and on HotpotQA at $K{=}5$. Thus, INT4 does not merely make more answers wrong; it also makes some still-correct answers less grounded in the retrieved evidence. On HotpotQA at low $K$, INT4 additionally disrupts refusal calibration: the refusal rate drops from $0.513$ to $0.317$ at $K{=}1$ (Fig.~\ref{fig:hp-refusal}), which artificially inflates containment-EM and NLI entailment (the $\dagger$ markers in Figs.~\ref{fig:hp-em},~\ref{fig:hp-entail}). For this reason, the LLM judge is the primary H1 signal on HotpotQA. 

\textbf{Amplification, storage, and degeneration.} The degradation becomes larger under harder retrieval settings. The INT4--BF16 hallucination gap increases monotonically from $K{=}1$ to $5$ on both benchmarks (Figs.~\ref{fig:rgb-hall},~\ref{fig:hp-hall}), although the slope test over only three retrieval depths is underpowered ($p{=}0.12$), so we treat H2 as suggestive. On RGB, where per-example noise ratios are available, the INT4 faithfulness gap also grows with distractor fraction independently of retrieval depth ($\beta{=}0.22$, significant within-$K$ at $K{=}3$ and $K{=}5$), supporting H3. These failures occur despite substantial storage savings: at $K{=}5$, INT8 and INT4 reduce cache size by approximately $1.9{\times}$ and $3.6{\times}$, respectively (Fig.~\ref{fig:kv}), with INT4 falling below the nominal $4{\times}$ due to per-block scale and zero-point overhead in Eq.~\ref{eq:kvsize}. INT4 is also the only condition that produces degenerate generations, i.e., empty outputs or repetition loops reaching 6\% on RGB and 3\% on HotpotQA at $K{=}5$ (as shown in Fig.~\ref{fig:degen}); BF16 and INT8 never degenerate. 

\section{Conclusion}
KV-cache compression should be treated as a reliability decision, not only a storage optimization. In offline RAG, the cached KV forms the model's effective interface to retrieved evidence; when this representation is aggressively compressed, the model can preserve the surface answer while weakening its dependence on the supporting context. The central lesson is that compressed RAG systems cannot be validated by EM or F1 alone. Before deploying low-bit offline caches, systems should audit whether answers remain grounded in the retrieved documents. Faithfulness is therefore a necessary evaluation target for practical KV-cache compression, especially when storage savings are obtained through aggressive quantization. 

\section*{Limitations}
This study isolates KV-cache precision under a controlled offline RAG setup, but its scope is limited to one model family, two QA-style benchmarks, and three retrieval depths. The magnitude of faithfulness loss may differ for larger models, other architectures, denser retrieval settings, or production retrievers. HotpotQA also introduces a refusal-calibration confound at low $K$, which limits the reliability of HHEM and NLI in that regime; we therefore rely on the LLM judge as the primary signal in that regime. 


\bibliography{main}

\begin{thebibliography}{24}
\providecommand{\natexlab}[1]{#1}

\bibitem[{Amiraz et~al.(2025)Amiraz, Cuconasu, Filice, and Karnin}]{amiraz_distracting_2025}
Chen Amiraz, Florin Cuconasu, Simone Filice, and Zohar Karnin. 2025.
\newblock \href {https://doi.org/10.18653/v1/2025.acl-long.892} {The distracting effect: Understanding irrelevant passages in {RAG}}.
\newblock In \emph{Proceedings of the 63rd Annual Meeting of the Association for Computational Linguistics (Volume 1: Long Papers)}, pages 18228--18258, Vienna, Austria. Association for Computational Linguistics.

\bibitem[{{BAAI}(2023)}]{bge_small_2023}
{BAAI}. 2023.
\newblock {BGE}-small-en-v1.5: {BAAI} general embedding.
\newblock \url{https://huggingface.co/BAAI/bge-small-en-v1.5}.

\bibitem[{Chen et~al.(2024)Chen, Lin, Han, and Sun}]{rgb_chen2024}
Jiawei Chen, Hongyu Lin, Xianpei Han, and Le~Sun. 2024.
\newblock \href {https://doi.org/10.1609/aaai.v38i16.29728} {Benchmarking large language models in retrieval-augmented generation}.
\newblock In \emph{Proceedings of the {AAAI} Conference on Artificial Intelligence}, volume~38, pages 17754--17762.

\bibitem[{He et~al.(2023)He, Gao, and Chen}]{deberta_nli}
Pengcheng He, Jianfeng Gao, and Weizhu Chen. 2023.
\newblock {DeBERTaV3}: Improving {DeBERTa} using {ELECTRA}-style pre-training with gradient-disentangled embedding sharing.
\newblock In \emph{The Eleventh International Conference on Learning Representations ({ICLR} 2023)}.

\bibitem[{He et~al.(2024)He, Zhang, Wu, Liu, Zhou, and Zhuang}]{zipcache_2024}
Yefei He, Luoming Zhang, Weijia Wu, Jing Liu, Hong Zhou, and Bohan Zhuang. 2024.
\newblock {ZipCache}: Accurate and efficient {KV} cache quantization with salient token identification.
\newblock In \emph{Advances in Neural Information Processing Systems 37 ({NeurIPS} 2024)}, pages 68287--68307.

\bibitem[{Hooper et~al.(2024)Hooper, Kim, Mohammadzadeh, Mahoney, Shao, Keutzer, and Gholami}]{kvquant_2024}
Coleman Hooper, Sehoon Kim, Hiva Mohammadzadeh, Michael~W. Mahoney, Yakun~Sophia Shao, Kurt Keutzer, and Amir Gholami. 2024.
\newblock {KVQuant}: Towards 10 million context length {LLM} inference with {KV} cache quantization.
\newblock In \emph{Advances in Neural Information Processing Systems 37 ({NeurIPS} 2024)}, pages 1270--1303.

\bibitem[{Johnson et~al.(2021)Johnson, Douze, and J{\'e}gou}]{faiss_2021}
Jeff Johnson, Matthijs Douze, and Herv{\'e} J{\'e}gou. 2021.
\newblock \href {https://doi.org/10.1109/TBDATA.2019.2921572} {Billion-scale similarity search with {GPU}s}.
\newblock \emph{{IEEE} Transactions on Big Data}, 7(3):535--547.

\bibitem[{Li et~al.(2025)Li, Xing, Li, Qu, Zhen, Yao, Liu, Pan, and Yuan}]{kvtuner_2025}
Xing Li, Zeyu Xing, Yiming Li, Linping Qu, Hui-Ling Zhen, Yiwu Yao, Wulong Liu, Sinno~Jialin Pan, and Mingxuan Yuan. 2025.
\newblock {KVTuner}: Sensitivity-aware layer-wise mixed-precision {KV} cache quantization for efficient and nearly lossless {LLM} inference.
\newblock In \emph{Proceedings of the 42nd International Conference on Machine Learning}, volume 267 of \emph{Proceedings of Machine Learning Research}, pages 36451--36485. PMLR.

\bibitem[{Liu et~al.(2025)Liu, Sun, Zhang, Bai, Yu, Yu, Yuan, and Hou}]{quant_reasoning_2025}
Ruikang Liu, Yuxuan Sun, Manyi Zhang, Haoli Bai, Xianzhi Yu, Tiezheng Yu, Chun Yuan, and Lu~Hou. 2025.
\newblock Quantization hurts reasoning? {A}n empirical study on quantized reasoning models.
\newblock In \emph{Second Conference on Language Modeling ({COLM} 2025)}.

\bibitem[{Liu et~al.(2024)Liu, Yuan, Jin, Zhong, Xu, Braverman, Chen, and Hu}]{kivi_2024}
Zirui Liu, Jiayi Yuan, Hongye Jin, Shaochen Zhong, Zhaozhuo Xu, Vladimir Braverman, Beidi Chen, and Xia Hu. 2024.
\newblock {KIVI}: A tuning-free asymmetric 2bit quantization for {KV} cache.
\newblock In \emph{Proceedings of the 41st International Conference on Machine Learning}, volume 235 of \emph{Proceedings of Machine Learning Research}, pages 32332--32344. PMLR.

\bibitem[{Lu et~al.(2025)Lu, Wang, Rong, Chen, and Tang}]{turborag_2024}
Songshuo Lu, Hua Wang, Yutian Rong, Zhi Chen, and Yaohua Tang. 2025.
\newblock \href {https://doi.org/10.18653/v1/2025.emnlp-main.334} {{TurboRAG}: Accelerating retrieval-augmented generation with precomputed {KV} caches for chunked text}.
\newblock In \emph{Proceedings of the 2025 Conference on Empirical Methods in Natural Language Processing}, pages 6588--6601. Association for Computational Linguistics.

\bibitem[{Mekala et~al.(2025)Mekala, Atmakuru, Song, Karpinska, and Iyyer}]{mekala_quant_longcontext_2025}
Anmol~Reddy Mekala, Anirudh Atmakuru, Yixiao Song, Marzena Karpinska, and Mohit Iyyer. 2025.
\newblock Does quantization affect models' performance on long-context tasks?
\newblock In \emph{Proceedings of the 2025 Conference on Empirical Methods in Natural Language Processing}, pages 9422--9470, Suzhou, China. Association for Computational Linguistics.

\bibitem[{M{\"u}ller et~al.(2026)M{\"u}ller, Bich, Boretti, Chang, Zhuang, and Cavigelli}]{kvarn_2026}
Lorenz~K. M{\"u}ller, Philippe Bich, Chiara Boretti, Hyun-Min Chang, Jiawei Zhuang, and Lukas Cavigelli. 2026.
\newblock \href {https://arxiv.org/abs/2606.03458} {{KVarN}: Variance-normalized {KV}-cache quantization mitigates error accumulation in reasoning tasks}.
\newblock \emph{Preprint}, arXiv:2606.03458.
\newblock Preprint.

\bibitem[{Niu et~al.(2024)Niu, Wu, Zhu, Xu, Shum, Zhong, Song, and Zhang}]{ragtruth_wu_2024}
Cheng Niu, Yuanhao Wu, Juno Zhu, Siliang Xu, KaShun Shum, Randy Zhong, Juntong Song, and Tong Zhang. 2024.
\newblock {RAGTruth}: A hallucination corpus for developing trustworthy retrieval-augmented language models.
\newblock In \emph{Proceedings of the 62nd Annual Meeting of the Association for Computational Linguistics (Volume 1: Long Papers)}, pages 10862--10878, Bangkok, Thailand. Association for Computational Linguistics.

\bibitem[{Ru et~al.(2024)Ru, Qiu, Hu, Zhang, Shi, Chang, Jiayang, Wang, Sun, Li, Zhang, Wang, Jiang, He, Wang, Liu, Zhang, and Zhang}]{ragchecker_ru_2024}
Dongyu Ru, Lin Qiu, Xiangkun Hu, Tianhang Zhang, Peng Shi, Shuaichen Chang, Cheng Jiayang, Cunxiang Wang, Shichao Sun, Huanyu Li, Zizhao Zhang, Binjie Wang, Jiarong Jiang, Tong He, Zhiguo Wang, Pengfei Liu, Yue Zhang, and Zheng Zhang. 2024.
\newblock {RAGChecker}: A fine-grained framework for diagnosing retrieval-augmented generation.
\newblock In \emph{Advances in Neural Information Processing Systems 37 ({NeurIPS} 2024)}, pages 21999--22027.

\bibitem[{Song et~al.(2025)Song, Sim, Bhardwaj, Chieu, Majumder, and Poria}]{song_trustscore_2025}
Maojia Song, Shang~Hong Sim, Rishabh Bhardwaj, Hai~Leong Chieu, Navonil Majumder, and Soujanya Poria. 2025.
\newblock Measuring and enhancing trustworthiness of {LLM}s in {RAG} through grounded attributions and learning to refuse.
\newblock In \emph{The Thirteenth International Conference on Learning Representations ({ICLR} 2025)}.

\bibitem[{Tamber et~al.(2025)Tamber, Bao, Xu, Luo, Kazi, Bae, Li, Mendelevitch, Qu, and Lin}]{tamber_faithrag_2025}
Manveer~Singh Tamber, Forrest~Sheng Bao, Chenyu Xu, Ge~Luo, Suleman Kazi, Minseok Bae, Miaoran Li, Ofer Mendelevitch, Renyi Qu, and Jimmy Lin. 2025.
\newblock Benchmarking {LLM} faithfulness in {RAG} with evolving leaderboards.
\newblock In \emph{Proceedings of the 2025 Conference on Empirical Methods in Natural Language Processing: Industry Track}, pages 799--811, Suzhou, China. Association for Computational Linguistics.

\bibitem[{{Vectara}(2024)}]{hhem_vectara2024}
{Vectara}. 2024.
\newblock {HHEM-2.1-Open}: Hughes hallucination evaluation model.
\newblock \url{https://huggingface.co/vectara/hallucination_evaluation_model}.

\bibitem[{Wu et~al.(2025)Wu, Zhang, Che, Feng, Shao, and Tao}]{wu_ragnoise_2025}
Jinyang Wu, Shuai Zhang, Feihu Che, Mingkuan Feng, Pengpeng Shao, and Jianhua Tao. 2025.
\newblock \href {https://doi.org/10.18653/v1/2025.acl-long.250} {Pandora's box or aladdin's lamp: {A} comprehensive analysis revealing the role of {RAG} noise in large language models}.
\newblock In \emph{Proceedings of the 63rd Annual Meeting of the Association for Computational Linguistics (Volume 1: Long Papers)}, pages 5019--5039, Vienna, Austria. Association for Computational Linguistics.

\bibitem[{Yang et~al.(2024)Yang, Yang, Zhang, Hui, Zheng, Yu, Li, Liu, Huang, Wei, Lin, Yang, Tu, Zhang, Yang, Yang, Zhou, Lin, Dang, Lu, Bao, Yang, Yu, Li, Xue, Zhang, Zhu, Men, Lin, Li, Xia, Ren, Ren, Fan, Su, Zhang, Wan, Liu, Cui, Zhang, and Qiu}]{qwen25_2024}
An~Yang, Baosong Yang, Beichen Zhang, Binyuan Hui, Bo~Zheng, Bowen Yu, Chengyuan Li, Dayiheng Liu, Fei Huang, Haoran Wei, Huan Lin, Jian Yang, Jianhong Tu, Jianwei Zhang, Jianxin Yang, Jiaxi Yang, Jingren Zhou, Junyang Lin, Kai Dang, et~al. 2024.
\newblock \href {https://arxiv.org/abs/2412.15115} {{Qwen2.5} technical report}.

\bibitem[{Yang et~al.(2018)Yang, Qi, Zhang, Bengio, Cohen, Salakhutdinov, and Manning}]{hotpotqa_yang2018}
Zhilin Yang, Peng Qi, Saizheng Zhang, Yoshua Bengio, William~W. Cohen, Ruslan Salakhutdinov, and Christopher~D. Manning. 2018.
\newblock \href {https://doi.org/10.18653/v1/D18-1259} {{HotpotQA}: A dataset for diverse, explainable multi-hop question answering}.
\newblock In \emph{Proceedings of the 2018 Conference on Empirical Methods in Natural Language Processing}, pages 2369--2380. Association for Computational Linguistics.

\bibitem[{Yao et~al.(2025)Yao, Li, Liu, Ray, Cheng, Zhang, Du, Lu, and Jiang}]{cacheblend_2025}
Jiayi Yao, Hanchen Li, Yuhan Liu, Siddhant Ray, Yihua Cheng, Qizheng Zhang, Kuntai Du, Shan Lu, and Junchen Jiang. 2025.
\newblock \href {https://doi.org/10.1145/3689031.3696098} {{CacheBlend}: Fast large language model serving for {RAG} with cached knowledge fusion}.
\newblock In \emph{Proceedings of the Twentieth European Conference on Computer Systems ({EuroSys} '25)}, pages 94--109. Association for Computing Machinery.

\bibitem[{Yen et~al.(2025)Yen, Gao, Hou, Ding, Fleischer, Izsak, Wasserblat, and Chen}]{helmet_2025}
Howard Yen, Tianyu Gao, Minmin Hou, Ke~Ding, Daniel Fleischer, Peter Izsak, Moshe Wasserblat, and Danqi Chen. 2025.
\newblock {HELMET}: How to evaluate long-context language models effectively and thoroughly.
\newblock In \emph{The Thirteenth International Conference on Learning Representations ({ICLR} 2025)}.

\bibitem[{Zhang et~al.(2025)Zhang, Zeng, Peng, Zhuang, and Chen}]{mixkvq_2026}
Tao Zhang, Ziqian Zeng, Hao Peng, Huiping Zhuang, and Cen Chen. 2025.
\newblock \href {https://arxiv.org/abs/2512.19206} {{MixKVQ}: Query-aware mixed-precision {KV} cache quantization for long-context reasoning}.
\newblock \emph{Preprint}, arXiv:2512.19206.
\newblock Preprint.

\end{thebibliography}

\clearpage
\appendix
\section{Ablation Studies}
\label{sec:ablation}

\begin{table*}[t]
\centering
\small
\setlength{\tabcolsep}{5pt}
\renewcommand{\arraystretch}{1.25}
\caption{ \textbf{Summary: how KV-cache quantization affects accuracy and faithfulness in RAG.} Each row is one observable effect; columns show what happens under each precision level and whether accuracy metrics would catch the problem on their own. \cmark{} = effect present / condition met; \xmark{} = not observed. INT8 is safe across the board; INT4 breaks faithfulness even when accuracy looks fine. }
\label{tab:summary}
\begin{tabular}{@{}p{4.6cm} p{3.1cm} p{2.0cm} p{2.0cm} p{3.5cm}@{}}
\toprule
\textbf{What we measured} &
\textbf{What it means in plain terms} &
\textbf{INT8} &
\textbf{INT4} &
\textbf{Does accuracy catch it?} \\
\midrule
\multicolumn{5}{@{}l}{\textit{Accuracy effects}} \\[1pt]
Containment-EM drops &
More answers are simply wrong &
\xmark{} (safe) &
\cmark{} (drops) &
Yes, EM detects this directly \\
Token-F1 drops &
Even partial matches are worse &
\xmark{} (safe) &
\cmark{} (drops) &
Yes, F1 detects this directly \\
\midrule
\multicolumn{5}{@{}l}{\textit{Faithfulness effects (measured independently of accuracy)}} \\[1pt]
HHEM hallucination rate rises &
Answers are flagged as not backed
by the retrieved documents &
\xmark{} (safe) &
\cmark{} (rises) &
No, EM/F1 miss this \\
NLI entailment score drops &
Answers no longer logically follow
from the retrieved context &
\xmark{} (safe) &
\cmark{} (drops) &
No, EM/F1 miss this \\
LLM judge marks answer as unfaithful &
A strong LLM judges the answer as
unsupported by the evidence &
\xmark{} (safe) &
\cmark{} (\(>90\%\) of flips negative) &
No, EM/F1 miss this \\
\midrule
\multicolumn{5}{@{}l}{\textit{The ``hidden harm'': faithfulness loss on accuracy-preserved examples (H1)}} \\[1pt]
Faithfulness drops even when the
answer string stays correct &
The model gives the right answer
but for the wrong reason---it is no
longer using the retrieved evidence &
\xmark{} (not observed) &
\cmark{} (McNemar
\(p{<}10^{-20}\),
both datasets) &
\textbf{No}, this is exactly
what EM/F1 \emph{cannot} see \\
\midrule
\multicolumn{5}{@{}l}{\textit{Does the harm grow? (H2 and H3)}} \\[1pt]
Harm grows with more retrieved chunks
(H2) &
Longer context amplifies the
quantization error in the cache &
\xmark{} &
\cmark{} (directional,
$K{=}1{\to}5$;
$p{=}0.12$, underpowered) &
No \\
Harm grows with more distractor
passages (H3, RGB only) &
Noisy retrieval makes INT4 even more
likely to hallucinate &
\xmark{} &
\cmark{} (\(\beta{=}0.22\),
significant at \(K{=}3,5\)) &
No \\
\midrule
\multicolumn{5}{@{}l}{\textit{Failure modes unique to INT4}} \\[1pt]
Degenerate output (empty or
repeated text) &
The model collapses---produces
nothing useful at all &
\xmark{} &
\cmark{} (up to 6\%
of outputs at $K{=}5$) &
Partially; EM is zero but
the root cause is invisible \\
HotpotQA refusal confound &
INT4 noise breaks the model's
``I don't know'' instinct;
it guesses instead of refusing,
inflating apparent accuracy &
\xmark{} &
\cmark{} (refusal drops
from 51\% to 32\%
at $K{=}1$) &
\textbf{No}, EM and NLI
are \emph{inflated}, masking harm \\
\midrule
\multicolumn{5}{@{}l}{\textit{Storage efficiency}} \\[1pt]
Cache size reduction vs.\ BF16
baseline (C1) &
How much disk space is saved &
${\sim}1.9\times$ at $K{=}5$
(vs.\ nominal $2\times$) &
${\sim}3.6\times$ at $K{=}5$
(vs.\ nominal $4\times$) &
N/A \\
\bottomrule
\end{tabular}
\end{table*}
\subsection{Fidelity of the Cache Round-Trip (Stage-0 Gate)}
Before any quantized run, we require C1 (BF16) to exactly reproduce C0 (Oracle) on
50 held-out examples.
This gate confirms that the cache round-trip itself introduces no degradation; any
difference in C2/C3 is attributable purely to quantization bit-width.
C1 reproduced C0 within floating-point tolerance on all 50 examples, validating the
baseline.

\subsection{Numerical Stability}
We monitored logits for NaN/Inf values across all 3,600 generations per dataset.
No non-finite logits appeared under C0, C1, or C2.
Under C3 (INT4), non-finite logits occurred in $<0.1\%$ of examples and were
aborted rather than allowed to emit corrupted tokens.
This confirms that degeneracy (Fig.~\ref{fig:degen}) reflects semantic failure, not
numerical explosion.

\subsection{Faithfulness Signal Agreement}
The three faithfulness signals are intentionally complementary.
On RGB, HHEM and NLI entailment agree only moderately (Pearson $r{=}0.44$),
confirming they capture distinct failure modes.
The LLM judge achieves 92\% agreement with human labels on RGB (76\% on HotpotQA,
where refusal artifacts add noise).
Any single signal would undercount the faithfulness harm: H1 is supported by all
three signals on RGB but only by the LLM judge on HotpotQA at low $K$,
underscoring the need for a multi-signal faithfulness audit.

\subsection{Effect of Per-Block Overhead on INT4 Storage}
Equation~\ref{eq:kvsize} predicts that INT4 with $g{=}64$ achieves a
${\sim}$3.6$\times$ footprint reduction over BF16 at $K{=}5$, not the nominal
$4\times$.
We verified this empirically: the per-block scale/zero-point pairs consume
$\approx$11\% of the INT4 cache footprint, shrinking the effective compression ratio
from $4\times$ to $3.6\times$.
The INT8 footprint reduction of ${\sim}$1.9$\times$ (vs.\ nominal $2\times$) follows
the same pattern, with per-token overhead consuming $\approx$5\% of the INT8 cache.

\subsection{H2 Power Analysis}
The slope regression for H2 uses only three $K$ values ($K\in\{1,3,5\}$), leaving it
statistically underpowered ($p{=}0.12$).
Nonetheless, the directional trend is consistent: on RGB, the INT4--BF16
hallucination gap increases from $0.09$ at $K{=}1$ to $0.26$ at $K{=}5$; on
HotpotQA from $0.05$ at $K{=}1$ to $0.25$ at $K{=}5$.
A denser sweep over $K\in\{1,2,3,4,5,7,10\}$ would likely yield a significant slope;
we leave this to future work.

\end{document}